%% file: main.tex
\documentclass{article}
\usepackage{spconf,amsmath,graphicx}

\usepackage{microtype}
\usepackage{graphicx}
\usepackage{subfig}
\usepackage{booktabs} 
\usepackage{xtab}
\usepackage{multirow,multicol}
\usepackage{threeparttable}
\usepackage{tablefootnote}
\usepackage{makecell}
\usepackage{algpseudocode}
\usepackage{amsmath}
\usepackage{amsfonts}
\usepackage{amssymb}
\usepackage{amsthm}
\usepackage{mathtools}

\usepackage{soul}

\usepackage{pifont}
\usepackage{comment}
\usepackage{array}
\usepackage{colortbl}
\usepackage{url}
\usepackage{color,soul}
\usepackage[dvipsnames]{xcolor}
\usepackage{graphics}
\usepackage{wrapfig}
\usepackage{algorithm}

\usepackage{soul}
\usepackage{algorithm}
\usepackage{float}
\usepackage{tablefootnote}
\usepackage{multirow}
\usepackage[square,numbers,compress]{natbib}
\usepackage{pifont}
\newcommand\mynuma[1]{\ifcase#1 \or \ding{172}\or \ding{173}\or
  \ding{174}\or \ding{175}\or \ding{176}\or \ding{177}%
  \or \ding{178}\or \ding{179}\or \ding{180}\or \ding{181}\else *\fi\relax}
\newcommand\mynumb[1]{\ifcase#1 \or \ding{182}\or \ding{183}\or
  \ding{184}\or \ding{185}\or \ding{186}\or \ding{187}%
  \or \ding{188}\or \ding{189}\or \ding{190}\or \ding{191}\else *\fi\relax}

\newcommand{\shortarrow}[1][4.5pt]{\mathrel{%
   \hbox{\rule[\dimexpr\fontdimen22\textfont2-.2pt\relax]{#1}{.4pt}}%
   \mkern-4mu\hbox{\usefont{U}{lasy}{m}{n}\symbol{41}}}}

\title{\textit{DNN-Chip Predictor}: An Analytical Performance Predictor for DNN Accelerators with Various Dataflows and Hardware Architectures}
\name{Yang Zhao, Chaojian Li, Yue Wang, Pengfei Xu, Yongan Zhang, and Yingyan Lin}
\address{Department of Electrical and Computer Engineering, Rice University}
\begin{document}
%
\maketitle
\input{tex/0-Abstract.tex}

\input{tex/1-Introduction.tex}
\input{tex/2-Related-works.tex}

\input{tex/3-Proposed-revision.tex}
\input{tex/4-Experiment-revision.tex}

\input{tex/5-Conclusion.tex}

\bibliographystyle{IEEEbib}


\end{document}

%% file: tex/0-Abstract.tex
\begin{abstract}
\vspace{-0.5em}

The recent breakthroughs in deep neural networks (DNNs) have spurred a tremendously increased demand for DNN accelerators. However, designing DNN accelerators is non-trivial as it often takes months/years and requires cross-disciplinary knowledge. To enable fast and effective DNN accelerator development, we propose \textit{DNN-Chip Predictor}, an analytical performance predictor which can accurately predict DNN accelerators' energy, throughput, and latency prior to their actual implementation. Our \textit{Predictor} features two highlights: 
(1) its analytical performance formulation of DNN ASIC/FPGA accelerators facilitates fast design space exploration and optimization; and (2) it supports DNN accelerators with different algorithm-to-hardware mapping methods (i.e., dataflows) and hardware architectures. Experiment results based on 2 DNN models and 3 different ASIC/FPGA implementations show that our \textit{DNN-Chip Predictor}'s predicted performance differs from those of chip measurements of FPGA/ASIC implementation by no more than 17.66\% when using different DNN models, hardware architectures, and dataflows. 
The codes are available at \url{https://github.com/RICE-EIC/DNN-Chip-Predictor}.

\end{abstract}
\vspace{-0.3em}
\begin{keywords}
DNN accelerator, ASIC, FPGA, design simulator, design automation
\end{keywords}

%% file: tex/1-Introduction.tex
 \vspace{-0.9em}
\section{Introduction}\label{sec:intro}
\vspace{-0.8em}
Deep Neural Networks (DNNs) have achieved record-breaking performance in various applications, such as image classification~\cite{VGG2014, wang2019e2} and natural language processing~\cite{xiong2017microsoft}. However, their powerful performance often comes with a prohibitive complexity \cite{liu2018adadeep,wang2018energynet,fracskip,deepkmeans,wang2019dual}. Moreover, DNN-based applications often require not only high accuracy, but also aggressive hardware performance, including high throughput, low latency, and high energy efficiency. As such, there has been intensive research on DNN accelerators in order to take advantage of different hardware platforms, such as FPGAs and ASICs, for improving DNN acceleration efficiency ~\cite{Ziyun-ISSC2019,Ziyun-JSSC2019,8050797,du2015shidiannao,eyeriss}.

While DNN accelerators can be 1000$\times$ more efficient than general purpose computing platforms~\cite{han2016eie}, developing DNN accelerators presents significant challenges, because: (1) mainstream DNNs have millions of parameters and billions of operations; (2) the design space of DNN accelerator is large due to numerous design choices of architectures, hardware IPs, DNN-to-accelerator-mappings, etc.; and (3) there is an algorithm/hardware co-design need for the same DNN functionality to have a different decomposition that would require different hardware IPs and thus correspond to dramatically different hardware performance/energy/area trade-offs. Therefore, high-quality DNN accelerators often take months/years to design and require a large team of cross-disciplinary experts with knowledge in DNN algorithms, micro-architectures, and physical chip design. Such a barrier makes it difficult to scientifically explore innovative DNN accelerator design and thus limits DNNs' more extensive applications.

To address the aforementioned challenges, we propose \textit{DNN-Chip Predictor}, an analytical performance predictor which can efficiently and accurately predict DNN accelerators' performance prior to time-consuming ASIC/FPGA hardware implementation. Specifically, our \textit{Predictor} formulates DNN accelerators' energy, throughput, and latency based on parameters that characterize the DNN models and corresponding accelerators' architectures and algorithm-to-hardware mapping methods (i.e., dataflows). Such a generic \textit{Predictor} (1) enables fast evaluation of DNN accelerator innovations and (2) can be used as an efficient design exploration and optimization tool for DNN accelerators, given their large design space.
To the best of our knowledge, our proposed \textit{Predictor} is \textbf{the first} that highlights the following three features simultaneously for practical and wide adoption: (1) analytical and thus fast; (2) covering both ASIC and FPGA DNN accelerators; (3) are validated using different DNN models and accelerator designs (i.e., architectures, dataflows, and process technologies). 

%% file: tex/2-Related-works.tex
\vspace{-1.2em}
\section{Background}\label{sec:background}
\vspace{-1em}
\textbf{DNN Accelerators.} There have been intensive studies of DNN accelerators. 
For example, the first well-optimized FPGA DNN accelerator~\cite{zhang2015optimizing} uses loop tiling; 
the DianNao series~\cite{du2015shidiannao, chen2014diannao} is an early effort on synthesis based ASIC accelerators; Eyeriss proposes a row-stationary dataflow~\cite{eyeriss} to reduce expensive DRAM accesses; and Google TPUs \cite{cloudtpu, edgetpu} use a systolic array to achieve high throughput. \\
\textbf{DNN Accelerator Performance Prediction.}
DNNs often feature a high complexity while there exists various opportunities for reuse, pipeline, and resource allocation to maximize DNN accelerators' performance. Therefore, an accurate yet fast performance predictor is desired to enable efficient design space exploration and optimization with different performance trade-offs.
Various methods have been developed for predicting or simulating DNN accelerators' performance.
Roofline models~\cite{zhang2015optimizing, tangmlpat} and customized analytical models which are closely tied to the specific design attributes
~\cite{zhang2018dnnbuilder, liu2017throughput, samajdar2018scale} are used. However, the roofline model lack fine-grained estimation and customized models are not general as desired. 
Timeloop~\cite{parashar2019timeloop} and Eyeriss~\cite{chen2018eyerissv2} use \textit{for} and \textit{parallel-for} to describe the temporal and spatial mapping of DNN accelerators. Specifically, Timeloop obtains the number of memory accesses and estimates the latency by calculating the maximum isolated execution cycle across all hardware IPs based on a double-buffering assumption. 
Accelergy~\cite{wu2019accelergy} proposes a configuration language to describe  hardware architectures and depends on plug-ins, e.g., Timeloop, to calculate the energy as in~\cite{eyeriss}.
The work in \cite{yang2018dnn} adopts Halide~\cite{ragan2013halide}, a domain-specific language for image processing applications, and proposes a modeling framework which is similar to that of ~\cite{eyeriss}. MAESTRO~\cite{kwon2019understanding} is the very first to adopt a data-centric approach.

%% file: tex/3-Proposed-revision.tex
\vspace{-1.3em}
\section{The Proposed \textit{DNN-Chip Predictor} }\label{sec:proposed}
    \vspace{-1em}
This section presents the proposed \textit{DNN-Chip Predictor} which is an analytical modelling framework to formulate DNN inference accelerators' energy cost, latency, and throughput when employing different dataflows and hardware architectures. We first introduce the employed design space description method, and then describe the developed performance models.
\textbf{The advantages} of the \textit{DNN-Chip Predictor} are that it (1) matches well with actual implementation results ($<$18\%); (2) is analytical and intuitive (directly ties to the DNN model and accelerator parameters), facilitating its ease of use for time-efficient design space exploration and optimization; and (3) is programmer friendly and compatible with commonly used DNN frameworks (e.g., Pytorch \cite{paszke2017automatic}) thanks to its adopted generic description of DNN accelerators' design space.

    \vspace{-1.1em}
\subsection{Design Space Description}
    \vspace{-0.6em}
    \label{subsec:design_space}
   
\begin{figure}[h]
   \vspace{-0.5em}
  \begin{algorithm}[H]
    \begin{algorithmic}[1]
    \Statex
    \State{\hspace{0pt}  {{\color{orange}// DRAM level}} }
    \State{\hspace{0pt}for ($m_3$=0; $m_3<M_3$; $m_3$++) \{}
    \State{\hspace{10pt}  {{\color{orange}// Global buffer level}} }
    \State{\hspace{10pt} {\color{blue}$\Rightarrow${$GB_{weight}$ refresh} }}
    \State{\hspace{10pt} {\color{blue} $\Rightarrow${$GB_{input}$ refresh} }}
    \State{\hspace{10pt}for ($e_2$=0; $e_2<E_2$; $e_2$++) \{}
    \State{\hspace{20pt} {\color{blue}$\Rightarrow${ $GB_{output}$ refresh} }}
    \State{\hspace{20pt}for ($c_2$=0; $c_2<C_2$; $c_2$++) \{}
    \State{\hspace{30pt} {{\color{orange}// NoC level}} }
    \State{\hspace{30pt}parallel-for ($f_1$=0; $f_1<F_1$; $f_1$++) \{}
    \State{\hspace{40pt} {{\color{orange}// RF level}} }
    \State{\hspace{40pt} {\color{blue}$\Rightarrow${$RF_{input}$ refresh, $RF_{weight}$ refresh}}}
    \State{\hspace{40pt}for ($r_0$=0; $r_0<R_0$; $r_0$++) \{}
    \State{\hspace{50pt}for ($s_0$=0; $s_0<S_0$; $s_0$++) \{}
    \State{\hspace{60pt}MAC operation}
    \State{\hspace{70pt} {\color{blue}$\Rightarrow${ $RF_{output}$ refresh} \}\}\}\}\}\}}}
    \end{algorithmic}
  \end{algorithm}
  \vspace{-2em}
  \caption{\textbf{A nested for-loop description of DNN accelerators' design space, using a CONV layer as an example, where {0,1,2,3} denotes the four memory hierarchies (i.e., RF, NoC, GB, and DRAM, respectively), and $M, C, R, S, E, F$ denote the six dimensions of a CONV layer (i.e., input/output channels, kernel width/height, and output feature map width/height, respectively).}}
  \label{fig:loop1}
   \vspace{-2em}
\end{figure}

For modeling DNN accelerators' performance given their large design space, one critical question is \textit{how to describe the whole design space, i.e., cover all possible design choices, in a way that is easy to follow?} For ease of use and better visualization, we adopt a nested for-loop description \cite{eyeriss} to describe the design space as shown in Fig.~\ref{fig:loop1}. Specifically, we employ (1) the primitive, \textit{for}, to describe the temporal operations of each process element (PE) as well as the temporal data tiling and mapping operations at the DRAM, global buffer (GB) and register file (RF) levels; and (2) the primitive, \textit{parallel-for}, to describe the spatial data tiling and mapping operations at the network-on-chip (NoC) level (i.e., in the PE array). Without loss of generality, we consider four levels of memory hierarchy, i.e., off-chip DRAM, on-chip GB, NoC in the PE array, and RF within the PEs. The design space of DNN accelerators mainly includes two aspects: hardware architectures and dataflows.

\textbf{Hardware architecture.} It can be described using a set of architecture-dependent hardware parameters and technology-dependent IP parameters. In particular, the architecture-dependent hardware parameters includes PE array architectures (e.g., spatial array, systolic array, and adder tree), number of PEs, NoC design (e.g., unicast, multicast, or broadcast), memory hierarchies, and the storage capacity and communication bandwidth of each memory hierarchy; the technology-dependent IP parameters includes unit energy/delay costs of (1) a MAC operation, (2) memory accesses to various memory hierarchies, and (3) the clock frequency.
\vspace{-0.1em}

\textbf{Dataflow.} This describes how a DNN is temporally and spatially scheduled to be executed in an accelerator. Specifically, a dataflow answers the following questions: (1) \textit{how to map and schedule the computations in the PE array and within each PE}?; and (2) \textit{what are the loop ordering and tiling factors on the DRAM and global buffer levels}? The former captures the design choice of holding a certain type of data locally in the PE once being fetched from the memories, e.g., row/weight/output stationary. 
The latter shows how to store data in SRAM and DRAM to accommodate data stationary effectively. These two questions can be described using
three groups of parameters as defined below in the context of the example in Fig.~\ref{fig:loop1}: \ul{Loop ordering factors} for the twenty-four nested for-loops associated with the six dimensions of the 3D convolution operation and the four considered memory hierarchies (i.e., DRAM, GB, NoC, and RF); \ul{Loop tiling factors} for the twenty-four nested  for-loops associated with the six dimensions of the 3D convolution operation and the four considered memory hierarchies; and \ul{Data access locations} in which of the nested for-loops we refresh the on-chip GB and in-PE RFs for the activations and weights. 

\subsection{The \textit{DNN-Chip Predictor}}
\vspace{-0.6em}
\label{subsec:framework}
\subsubsection{Overview}
\vspace{-0.6em}

\begin{figure}[!t]
       \vspace{-0.5em}
    \centerline{\includegraphics[width=75mm]{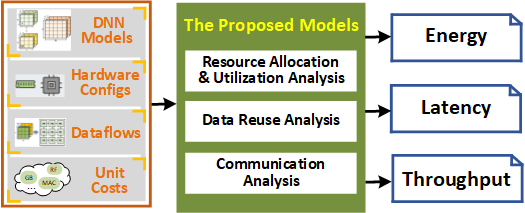}}
       \vspace{-0.8em}
    \caption{\textbf{A high-level view of the \textit{DNN-Chip Predictor}.} }
    \label{fig:df}
    \vspace{-1.5em}
\end{figure}

Fig.~\ref{fig:df} shows a high-level view of the proposed \textit{DNN-Chip Predictor}, which accepts DNN models (e.g., number of layers, layer structure, bit-precision, etc.), hardware architectures (e.g., memory hierarchy, number of PEs, NoC design, etc.), dataflows (e.g., row/weight/output stationary, loop tiling/ordering factors, etc.), and technology-dependent unit costs (e.g., unit energy/delay cost of a MAC operation and memory accesses to various memory hierarchies), and then outputs the estimated energy consumption, latency, and throughput when executing the DNN in a target accelerator. 
It thus can be used to (1) validate DNN accelerator techniques prior to the time- and cost-consuming DNN ASIC/FPGA accelerator implementation, and (2) perform time-efficient design space exploration and optimization. 


\vspace{-1.4em}
\subsubsection {The Proposed Analytical Models}
\vspace{-0.6em}
\label{subsec:perf_model}

This subsection introduces the \textit{Predictor}'s analytical models.
\vspace{0.2em}

\textbf{Energy Models}. 
DNN accelerators' energy cost include both computational ($E_{comp}$) and data movement ($E_{DM}$) costs, where $E_{comp} = N_{MAC}\times e_{MAC}$ with $N_{MAC}$ denoting the total number of MACs in the DNN. Similarly, 
the data movement cost can be calculated by multiplying the unit energy cost per access ($e_{DM_{i,j}}, j \in \{I, O, W\}$)  with the total number of accesses ($N_{DM_{i,j}}, j \in \{I, O, W\}$) to the $i$-th memory hierarchy (e.g., GB) using the $j$-th type of data (i.e., inputs ($I$), outputs ($O$), and weights ($W$)):  
\vspace{-0.5em}
\begin{normalsize}
\begin{flalign}
\label{eq:DM}
    &E_{DM} = \sum_{i \in S_{Memory}}\sum_{j \in \{I, O, W\}} N_{DM_{i,j}}, \times e_{DM_{i,j}}&
\end{flalign}
\end{normalsize}
\vspace{-1.2em}
\\where $S_{Memory} = \{DRAM \shortarrow GB, GB \shortarrow NoC, NoC \shortarrow RF, RF \shortarrow PE \}$ for inputs/weights; and $S_{Memory} = \{DRAM \leftrightarrow GB, GB \leftrightarrow NoC, NoC \leftrightarrow RF, RF \leftrightarrow MAC\}$ for outputs.
\vspace{0.2em}

\textbf{The key challenge is to obtain $N_{DM_{i,j}}$} for various memory hierarchies and data types when using different DNN models, hardware architectures, and dataflows. We are the first to find that $N_{DM_{i,j}}$ can be calculated as the product of the $j$-th data volume ($V_{ref_{i,j}}$)  involved in each refresh and the total number of such refreshes ($N_{ref_{i,j}}$) for the $i$-th memory: 
\vspace{-0.7em}
\begin{normalsize}
\begin{flalign}
    &N_{DM_{i,j}} = N_{ref_{i,j}} \times V_{ref_{i,j}}&
\end{flalign}
\end{normalsize}
\vspace{-1.4em}
\\To obtain $N_{ref_{i,j}}$ and $V_{ref_{i,j}}$, \textbf{we propose an intuitive methodology}: we first \ul{(1)} choose a refresh location, which can be straightforwardly decided once the dataflow is known, in the nested for-loops (see Fig. \ref{fig:loop1}) for a given  data type; \ul{(2)} $N_{ref_{i,j}}$ is equal to the product of all the loop bounds in the for-loops \textbf{above} the refresh location; and \ul{(3)} $V_{ref_{i,j}}$ is equal to the product of all the loop bounds in the for-loops \textbf{below} the refresh location and associated with the particular type of data. Once $N_{ref_{i,j}}$ and $V_{ref_{i,j}}$ are obtained, the energy can be calculated as: 
\vspace{-0.6em}
\begin{normalsize}
\begin{flalign}
    &E_{DRAM} = \sum_{j \in \{I,O,W\}} N_{ref_{GB,j}} \times V_{ref_{GB,j}} \times e_{DM_{DRAM,j}}&
\end{flalign}
\vspace{-3.em}
\begin{flalign}
    &E_{GB} = \sum_{j \in \{I,O,W\}} N_{ref_{RF,j}} \times V_{ref_{RF,j}}\times \frac{N_{PE}}{M_j} \times e_{DM_{GB,j}}&
\end{flalign}
\vspace{-3.em}
\begin{flalign}
    &E_{NoC} = \sum_{j \in \{I,O,W\}}N_{ref_{RF,j}} \times V_{ref_{RF,j}} \times N_{PE}\times e_{DM_{NoC,j}}&
\end{flalign}
\vspace{-2.5em}
\begin{flalign}
    &E_{RF} = \sum_{j \in \{I,O,W\}} N_{MAC} \times e_{DM_{RF,j}}&
\end{flalign}
\end{normalsize}
\vspace{-1.1em}
\\where $N_{PE}$ is the number of active PEs and $M_j$ is the number of PEs that share the same data.
\vspace{0.4em}

\textbf{Latency Models}. Similarly, the latency of DNN accelerators can be formulated as:
\vspace{-0.5em}
\begin{normalsize}
\begin{flalign}
    &L = L_{setup} + \max \{L_{DRAM}, L_{GB}, L_{comp}\}&
\end{flalign}
\end{normalsize}
\vspace{-1.6em}
\\where $L_{comp}$, $L_{DRAM}$, $L_{GB}$, and $L_{setup}$ denote the latency of computation in the PE array, accessing the DRAM from the GB, accessing the GB from an RF in the PEs, and setting up the first set of the weights and inputs, respectively. Adopting $N_{bit}^j$-bit precision for inputs/outputs/weights is $N_{bit}^j$, $j \in \{I,O,W\}$, we have:
\vspace{-4pt}
\begin{normalsize}
\begin{flalign}
    &L_{comp} = N_{MAC} \times t_{comp}&
\end{flalign}
\vspace{-2.5em}
\begin{flalign}
    &L_{DRAM} = \max_{j \in \{I,O,W\}} N_{ref_{GB,j}} \times \frac{V_{ref_{GB,j}}\times N_{bit}^j}{\min\{BW^j_{GB}, BW_{DRAM}\}}&
\end{flalign}
\vspace{-2.6em}
\begin{flalign}
    &L_{GB} = \max_{j \in \{I,O,W\}} N_{ref_{RF,j}} \times \frac{N_{ref_{RF,j}}\times N_{bit}^j\times N_{PE}}{BW^j_{GB}}&
\end{flalign}
\vspace{-2.4em}
\begin{flalign}
    &L_{setup} = \max (L^{'}_{DRAM}, L^{'}_{GB})&
\end{flalign}
\vspace{-2.6em}
\begin{flalign}
    &L^{'}_{DRAM} =\max_{j \in \{I,W\}} \frac{V_{ref_{GB,j}}\times N_{bit}^j}{\min\{BW_{j,{GB}}, BW_{DRAM}\}}&
\end{flalign}
\vspace{-2.em}
\begin{flalign}
    &L^{'}_{GB} =\max_{j \in \{I,W\}} \frac{N_{ref_{RF,j}}\times N_{bit}^j}{\min\{BW_{j,{RF}}, BW_{j,{GB}}\}}&
\end{flalign}
\end{normalsize}
\vspace{-1em}
\\where $BW^j_{i}$ is the memory bandwidth for the $i$-th memory hierarchy for the data type $j \in \{I,O,W\}$.

%% file: tex/4-Experiment-revision.tex
\vspace{-1em}
\section{Experiment Results}
\vspace{-1em}

We validate our proposed \textit{DNN-Chip Predictor} by comparing its predicted performance with actual chip measured ones in~\cite{eyeriss}, FPGA implementation results in~\cite{hao2019fpga}, and synthesis results based on a commercial CMOS technology, under the same experiment settings (e.g., unit energy, clock frequency, DNN model, architecture design and dataflow, etc). 

\begin{figure}[!t]
\vspace{-0.3em}
\captionsetup[subfigure]{labelformat=empty}
    \vspace{-1em}
    \centering
    \subfloat{{\includegraphics[width=40mm]{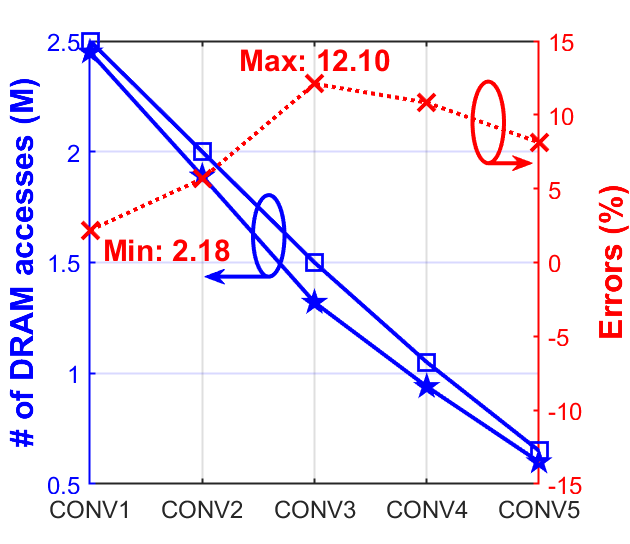} }}\hspace{1pt}
    \subfloat{{\includegraphics[width=40mm]{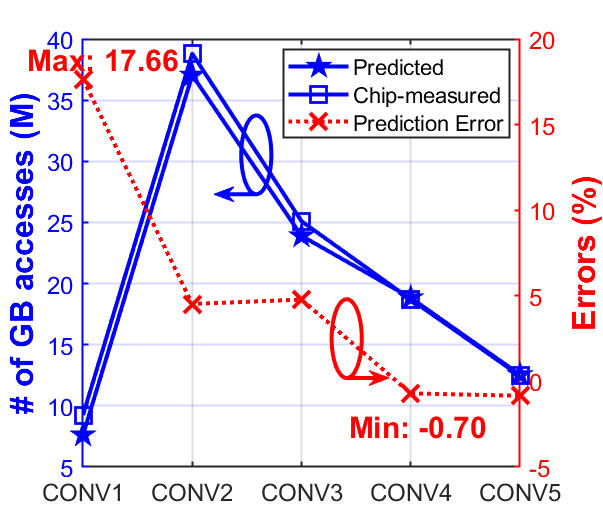} }}
    \vspace{-1em}
    \caption{\textbf{The \# of (L) DRAM and (R) GB accesses in Eyeriss~\cite{chen2017eyeriss} and our \textit{Predictor} for AlexNet's CONV layers.}}
    \vspace{-1.5em}
    \label{fig:val_2}
\end{figure}

\begin{table}[h]
\vspace{-0em}
\centering
\vspace{3mm}
\small
\setlength\tabcolsep{1pt}
\caption{\textbf{The energy breakdown from Eyeriss~\cite{chen2017eyeriss} and our \textit{Predictor}, for the CONV1 and CONV5 of AlexNet \cite{NIPS2012_4824}.}}
\vspace{-1.1em}
\begin{tabular}{@{}c | cc c cc c cc c cc@{}}\toprule
\multirow{2}{*}{Layer}& \multicolumn{2}{c}{comp.} & \phantom{x}& \multicolumn{2}{c}{RF} & \phantom{x}& \multicolumn{2}{c}{NoC} & \phantom{x}& \multicolumn{2}{c}{GB}\\ 
\cmidrule{2-3} \cmidrule{5-6} \cmidrule{8-9} \cmidrule{11-12}
& Meas. & Pred. && Meas. & Pred. && Meas. & Pred. && Meas. & Pred.  \\ 
\midrule
\multirow{1}{*}{CONV1}
$ $ & 16.7\% & 18.7\% && 79.6\% & 74.4\% && 1.7\% & 4.8\% && 2.0\% & 2.0\%\\
\multirow{1}{*}{\textbf{$\Delta$}}
$ $ & \multicolumn{2}{c}{\textbf{2.08\%}} & \phantom{x}& \multicolumn{2}{c}{\textbf{-5.15\%}} & \phantom{x}& \multicolumn{2}{c}{\textbf{3.10\%}} & \phantom{x}& \multicolumn{2}{c}{\textbf{-0.03\%}}\\
\midrule
\multirow{1}{*}{CONV5}
$ $ & 7.3\% & 7.5\% && 80.3\% & 79.1\% && 5.3\% & 7.0\% && 7.0\% & 6.3\%\\
\multirow{1}{*}{\textbf{$\Delta$}}
$ $ & \multicolumn{2}{c}{\textbf{0.26\%}} & \phantom{x}& \multicolumn{2}{c}{\textbf{-1.16\%}} & \phantom{x}& \multicolumn{2}{c}{\textbf{1.64\%}} & \phantom{x}& \multicolumn{2}{c}{\textbf{-0.74\%}}\\
\bottomrule
\end{tabular}
\label{fig:val_1}
\vspace{-1.5em}
\end{table}

\textbf{Validation against Chip Measurements.} For this set of experiments, we compare our \textit{Predictor}'s predicted performance with Eyeriss's chip measurement results using their normalized unit energy~\cite{eyeriss}. First, Table~\ref{fig:val_1} compares the \ul{energy breakdown} of AlexNet's first and fifth CONV layers (denoted as CONV1 and CONV5, respectively), showing that the maximum difference is 5.15\% and 1.64\%, respectively. 

Second, Fig.~\ref{fig:val_2} compares \ul{the number of DRAM/GB accesses}. The difference between the predicted number of DRAM accesses and Eyeriss's measured results is between 2.18\% and 12.10\%, while the difference in terms of GB accesses is between -0.70\% and 17.66\%. Our \textit{Predictor}'s predicted DRAM access number is smaller than that of Eyeriss because the RLC overhead of sparse activations depends on the input images and we lack the information about which set of images were used in Eyeriss's measurements. Additionally, Fig.~\ref{fig:val_2} shows that the difference between the predicted number of GB accesses and Eyeriss's results is less than 5\% except for the CONV1 layer where the relative larger prediction error is caused by its larger stride, which is 4. Specifically, a larger stride leads to lower utilization of inputs fetched from the GB, whereas our current \textit{Predictor} considers the generic case where stride is 1 as it is more often seen in recent DNN models. For better prediction accuracy, our \textit{Predictor} can be adjusted to cover cases with other stride values, i.e., more considered cases for the analytical models in Section \ref{subsec:perf_model}. 


\begin{figure}[!b]
  \vspace{-1.5em}
    \centerline{\includegraphics[width=85mm]{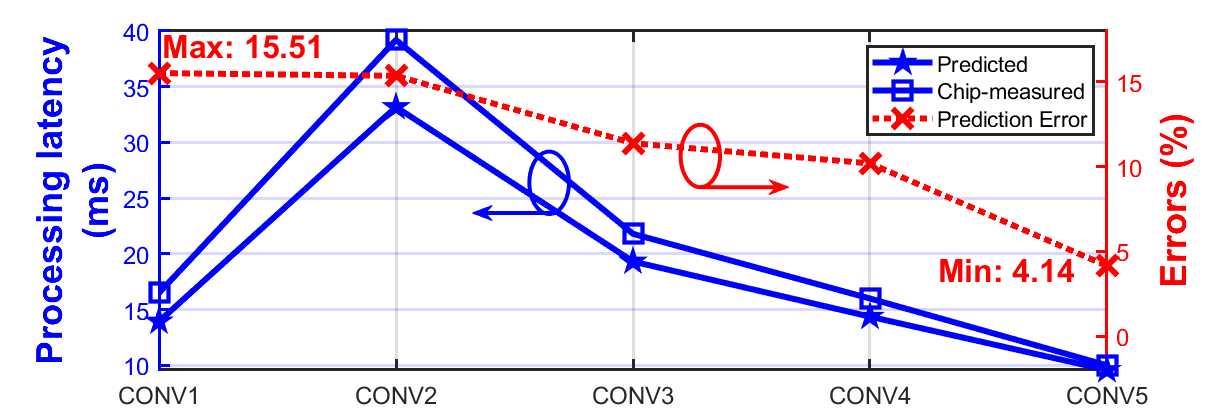}}
    \vspace{-1em}
    \caption{\textbf{Comparison on the inference latency from Eyeriss~\cite{chen2017eyeriss} and our \textit{Predictor} when running AlexNet.}}
    \vspace{-1em}
    \label{fig:latency}
\end{figure}

Third, Fig.~\ref{fig:latency} compares the \ul{latency} of executing AlexNet's five CONV layers, and shows that the predicted ones and Eyeriss's differ by $\le$ 15.51\%. The predicted latency is smaller than the measured one because our \textit{Predictor}'s analytical models do not consider the corner cycles when the memory accesses and computation can not be fully pipelined where processing stalls occur. Finally, the predicted \ul{throughput} of executing AlexNet is 46.0 GOPS while the one measured by Eyeriss is 51.6 GOPS, showing a prediction error of $\le$11\%. 

\begin{figure}[h]
   \vspace{-1.1em}
    \centerline{\includegraphics[width=88mm]{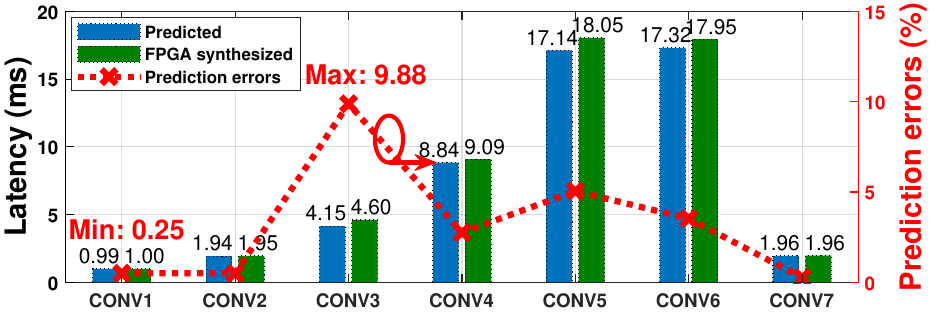}}
 
    \vspace{-0.9em}
    \caption{\textbf{Our \textit{Predictor}'s predicted latency and the FPGA measured one for the 7 CONV layers of SkyNet~\cite{hao2019fpga}.}}
    \vspace{-1em}
    \label{fig:val_fpga}
\end{figure}

\textbf{Validation against FPGA Implementation.} We compare our \textit{Predictor}'s predicted latency with FPGA measured ones under the same DNN model and hardware configurations~\cite{zhang2019skynet}. Specifically, for the FPGA one we use the open source implementation of the award winner~\cite{zhang2019skynet} in a state-of-the-art design contest~\cite{dac-contest}. Fig.~\ref{fig:val_fpga} shows that our \textit{Predictor}'s predicted latency differs from the FPGA-synthesized ones by $\le$ 16.84\%. Note that in FPGA implementations the GB can be partitioned into smaller chunks to be accessed simultaneously for increasing the parallelism and minimizing the latency. Our current models do not include the overhead of this partition, which is larger when the GB is partitioned into more chunks for layers with a larger size, leading to a larger prediction error for the CONV4/CONV5/CONV6 layers in Fig. ~\ref{fig:val_fpga}.

\textbf{Validation against Synthesis Results.} Table \ref{fig:val_syn} compares the \textit{Predictor}'s energy breakdown with that from the synthesis results for AlexNet's CONV3-CONV5 layers when using an in-house dedicated accelerator using a commercial 65nm CMOS technology. It can be seen from Table \ref{fig:val_syn} that the difference between our \textit{Predictor}'s predicted energy breakdown and that from the synthesis results is less than 5.28\%. 


\begin{table}
\vspace{-1.5em}
\centering
\vspace{3mm}
\small
\setlength\tabcolsep{2pt}
\caption{\textbf{The energy breakdown from synthesized results and our \textit{Predictor} for AlexNet's CONV3-CONV5 layers.}}
\vspace{-1.1em}
\begin{tabular}{@{}c | ccc c ccc c ccc@{}}\toprule
\multirow{2}{*}{Layer}& \multicolumn{3}{c}{comp. (\%)} & \phantom{x}& \multicolumn{3}{c}{RF(\%)} & \phantom{x}& \multicolumn{3}{c}{GB(\%)}\\ 
\cmidrule{2-4} \cmidrule{6-8} \cmidrule{10-12}
& Syn. & Pred. & \textbf{$\Delta$} && Syn. & Pred. & \textbf{$\Delta$} && Syn. & Pred. & \textbf{$\Delta$}  \\ 
\midrule
\multirow{1}{*}{CONV3}
$ $ & 38.76 & 34.49 &\textbf{4.26}  && 60.99 & 65.25 & \textbf{4.26} && 0.24 & 0.25 & \textbf{0.01} \\
\midrule
\multirow{1}{*}{CONV4}
$ $ & 39.46 & 34.28 &\textbf{5.19}  && 60.28 & 65.45 & \textbf{5.16} && 0.25 & 0.27 & \textbf{0.02} \\
\midrule
\multirow{1}{*}{CONV5}
$ $ & 31.13 & 25.85 &\textbf{5.28}  && 68.65 & 73.91 & \textbf{5.26} && 0.22 & 0.24 & \textbf{0.02} \\
\bottomrule
\end{tabular}
\label{fig:val_syn}
\vspace{-2em}
\end{table}

%% file: tex/5-Conclusion.tex
\vspace{-1.5em}
\section{Conclusion}
\vspace{-1.1em}
\label{sec:conc}
To close the gap between the growing demand for dedicated DNN accelerators with various specifications and the time-consuming and challenging DNN accelerator design, we develop \textit{DNN-Chip Predictor}, which can efficiently and effectively predict an accelerator's energy, latency, and resource consumption. Such an analytical performance prediction tool will facilitate fast development of innovations for not only DNN accelerators but also hardware-aware efficient DNNs. 

\vspace{-1em}



\section{Acknowledgement}
\vspace{-1em}
The work was supported in part by the National Science Foundation (NSF) awards  ECCS-1934767 and CCF-1838873. 